\documentclass{article}
\usepackage{iclr2026_conference,times}
\iclrfinalcopy

\usepackage{amsmath,amsfonts,bm}

\def\eqref#1{equation~\ref{#1}}

\def\1{\bm{1}}

\DeclareMathAlphabet{\mathsfit}{\encodingdefault}{\sfdefault}{m}{sl}
\SetMathAlphabet{\mathsfit}{bold}{\encodingdefault}{\sfdefault}{bx}{n}

\usepackage[utf8]{inputenc}
\usepackage[T1]{fontenc}
\usepackage{hyperref}
\usepackage{url}
\usepackage{booktabs}
\usepackage{amsfonts}
\usepackage{nicefrac}
\usepackage{microtype}
\usepackage{graphicx}
\usepackage{amsmath}
\usepackage{algorithm}
\usepackage{algorithmic}
\usepackage{geometry}
\usepackage{hyperref}
\usepackage{sidecap}
\usepackage{cleveref}

\title{PulseCX: Breaking the Closed-World Assumption in Real-Time CX}

\author{
  Rajat Agarwal \\
  Sprinklr AI, Gurugram, India \\
  \texttt{rajat.agarwal@sprinklr.com}
  \And
  Suvidha Tripathi \\
  Sprinklr AI, Gurugram, India \\
  \texttt{suvidha.tripathi@sprinklr.com} \\
  \And
  Shubham Sharma \\
  Sprinklr AI, Gurugram, India \\
  \texttt{shubham.sharma@sprinklr.com}
}

\begin{document}

\maketitle
\vspace{-5mm}

\begin{abstract}
Conversational AI agents in Customer Experience (CX) typically suffer from a \textbf{Closed-World Constraint}, ignoring high-velocity external shifts like viral trends or outages. Ad-hoc web search attempts to bridge this gap but often introduce prohibitive latency and \textit{context poisoning}. We introduce \textbf{PulseCX}, a framework that decouples knowledge acquisition from consumption. Adopting a \textbf{structure-first} paradigm, PulseCX employs an asynchronous agent to linearize signals into a \textbf{Decay-Aware Temporal Knowledge Graph (DA-TKG)} governed by reinforcement--decay dynamics to actively manage information lifecycles. By coupling this self-evolving memory with hierarchical intent gating, PulseCX removes synchronous search bottlenecks ($<$10ms overhead) and drives significant gains in Intent Resolution (IRR) and Customer Satisfaction (s-CSAT) in dynamic environments.
\end{abstract}

\vspace{-5mm}
\section{Introduction}
\vspace{-3mm}
Large Language Models (LLMs) have transformed automated Customer Experience (CX), enabling agents that reason and act across complex workflows. Yet most deployed agentic systems suffer from a fundamental disconnect: they are grounded in static internal artifacts (manuals, FAQs, dashboards) while operating in a world shaped by fast-moving external events. This gap is most visible during high-impact moments—such as influencer-driven PR crises or early-stage safety incidents (e.g., widespread reports of device overheating)—where customer perception shifts long before internal systems update. We define this mismatch as the \textbf{Closed-World Constraint}: the misalignment between an agent’s static internal state and high-velocity external reality. In practice, this manifests as \textit{context blindness} in \textbf{Dynamic intents}, where successful resolution depends on external, time-sensitive knowledge. Common attempts to bridge this gap via web search or news APIs frequently introduce unacceptable latency and severe \emph{context poisoning}, flooding the inference window with noisy, dangerous or contradictory signals which may become security risks \cite{chen2024agentpoison, eslami2025security}. 

To address this challenge, we propose \textbf{PulseCX}, a framework for selectively injecting real-time CXM context into agentic systems without sacrificing reliability. PulseCX adopts a \textbf{structure-first} paradigm~\citep{xu2025everything}, utilizing an \textbf{asynchronous discovery loop} to linearize raw external signals into structured Context Objects before they enter the reasoning window. Building on recent work on agent memory~\cite{hu2025memory}, we store these objects in a \textbf{Decay-Aware Temporal Knowledge Graph (DA-TKG)} governed by reinforcement--decay dynamics. This allows the system to actively maintain the lifecycle of facts—amplifying persistent signals (e.g., outages or fraud campaigns) while allowing transient noise to fade. Together, these mechanisms enable agents to resolve Dynamic intents while preserving low latency, bounded noise and poisoning signals , and reduced hallucination risk.
\vspace{-3mm}
\section{Related Work}
\vspace{-3mm}

\textbf{Agent Memory.} Prior work on memory for LLM-based agents emphasizes short-term, episodic, or long-term recall for coherence and planning, typically assuming a static external world~\cite{wang2024survey, hu2025memory}. Related efforts augment LLMs with temporal knowledge by injecting facts from predefined Temporal Knowledge Graphs into prompts or aligning model representations with TKG embeddings (e.g., TimeR4~\cite{chen2024temporal} and GenTKGQA~\cite{gao2024two}), but these approaches primarily target temporal reasoning for QA tasks rather than maintaining a lightweight, continuously evolving world-state memory for real-time agent grounding. PulseCX instead targets \emph{environment factual memory}, explicitly modeling the lifecycle of external events via reinforcement--decay dynamics. \textbf{Retrieval-Augmented and Online Agents.} Retrieval-Augmented Generation (RAG) and tool-using agents extend LLMs with external knowledge at inference time~\cite{lewis2020retrieval,yao2023react}. While effective for factual lookup, these approaches often incur latency and introduce noise in dynamic settings. PulseCX decouples context acquisition from inference by asynchronously structuring external signals into bounded, low-latency context. Related work in other ML systems have similarly explored gated mechanisms to control how contextual signals influence downstream optimization~\citep{gao2024ctxpipe}. \textbf{Context in CX Systems.} Most deployed CX systems rely on session history, customer profiles, and business rules. Although effective for routine interactions, they struggle with \emph{Dynamic intents} driven by rapidly evolving external events. While prior work has emphasized the importance of real-time context in CX systems~\cite{kim2022customer, motevalli2024enhancing, vashishth2024enhancing, badheka2014context}, such capabilities remain largely absent in practice. PulseCX addresses this gap by selectively injecting real-time situational context only when required.
Overall, PulseCX formalizes real-time external context as a first-class, evolving memory substrate rather than an ad-hoc retrieval artifact.
\vspace{-3mm}
\section{Methodology: The PulseCX Framework}
\label{sec:methodology}
\vspace{-3mm}

\begin{figure}[t] 
    \centering
    \vspace{-10pt} 
    \includegraphics[width=0.80\textwidth, trim={0 20pt 0 20pt}, clip]{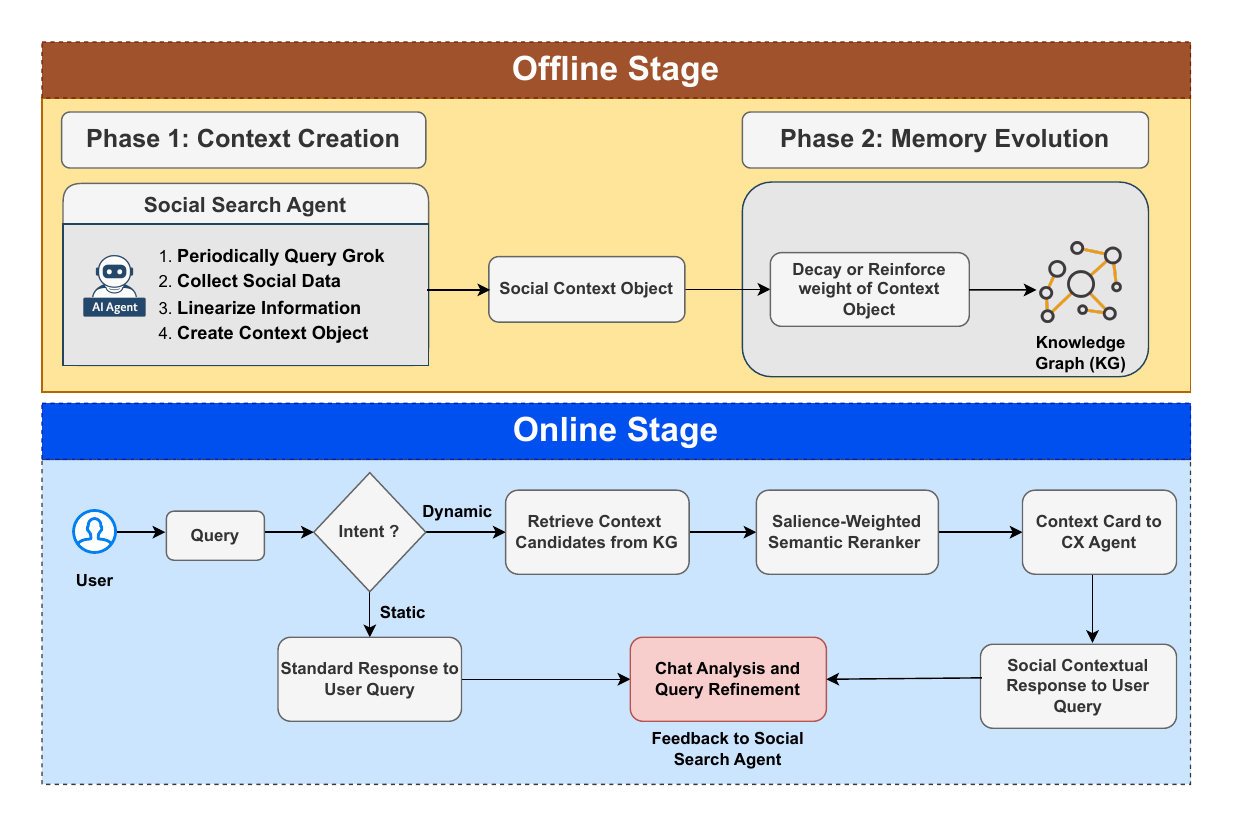}
    \vspace{-10pt} 
    \caption{\small \textbf{PulseCX Architecture:} \textbf{Offline Stage:} Asynchronous discovery and world modeling via DA-TKG memory evolution. \textbf{Online Stage:} Real-time intent-aware grounding and feedback-driven refinement}
    \label{fig:methodology}
    \vspace{-10pt} 
\end{figure}
PulseCX decouples high-latency knowledge acquisition from real-time inference via a two-stage architecture: \textbf{an Offline Asynchronous Stage} that structures world knowledge into a \textbf{Decay-Aware Temporal Knowledge Graph (DA-TKG)}, and an Online Synchronous Stage that selectively grounds user queries using this dynamic substrate, preserving contextual awareness of the "wild" without the latency or noise of synchronous web search.

\vspace{-3mm}
\subsection{Stage 1: Offline World Modeling (Structure-First)}
\label{sec:offline_stage}
\vspace{-2mm}
To capture high-velocity signals without slowing down inference, we employ a periodic \textbf{Social Search Agent (SSA)} that scans external sources at fixed intervals (e.g., $T=15m$). This stage transforms raw noise into structured intelligence through two phases.

\textbf{Phase 1: Linearization and Context Object Creation.}
Raw streams from web search or social APIs are unstructured and noisy. PulseCX enforces a \textbf{Structure-First} paradigm: the SSA output is immediately linearized into a structured \textbf{Context Object}. This object encapsulates not just the information, but its provenance and scope. By formalizing the input as a tuple $O = \{ \text{Type}, \text{Scope}, \text{Confidence}, \text{State} \}$, we ensure that only high-quality signals enter the graph (see Appendix \ref{app:context_object} for details).

\textbf{Phase 2: Memory Evolution (Reinforcement-Decay).} Unlike static databases, the DA-TKG is self-evolving, utilizing a \textbf{Reinforcement-Decay} mechanism to manage node lifecycles. To prune stale context (e.g., expired sales), node weight $W$ decays over time $t$:
\begin{equation}
    W_{t} = W_{last} \cdot e^{-\lambda(t - t_{last})} \quad \text{and} \quad W_{new} = \min(1.0, W_{decayed} + \beta \cdot C_{conf})
\end{equation}
where volatility coefficient $\lambda$ is high for viral trends and low for infrastructure persistence. Conversely, reinforcement constant $\beta$ ensures that signals re-confirmed by the SSA (with confidence $C_{conf}$) gain salience while one-off noise naturally fades. This ensures the graph accurately reflects the \textit{current} reality rather than historical noise.
\vspace{-3mm}
\subsection{Stage 2: Online Synchronous Grounding}
\label{sec:online_stage}
When a user queries the system, PulseCX executes a high-speed retrieval pipeline designed to inject relevant context without introducing latency or hallucinations. The process begins with a \textbf{Hierarchical Gating Engine} that classifies the user's intent. \textit{Static} intents (e.g., ``Reset Password'') bypass the graph entirely to prevent context poisoning, while \textit{Dynamic} intents (e.g., ``Why is the site slow?'') trigger retrieval. To select the most pertinent information, we employ a \textbf{Salience-Weighted Semantic Reranker} that scores candidate nodes ($N$) by combining their semantic similarity to the query ($Q$) with their current decay-adjusted weight ($W_t$):
\begin{equation}
    R(N) = \text{Sim}(Q, N) \cdot W_t
\end{equation}
This scoring function ensures the agent prioritizes nodes that are both relevant and temporally active, naturally filtering out expired events like old flash sales. The top-ranked nodes are then synthesized into a deterministic \textbf{Context Card}—a structured briefing injected into the main agent's system prompt (see Appendix \ref{app:context_cards}). Finally, the system closes the loop via a \textbf{Query Refiner} that analyzes chat logs for information gaps (e.g., unknown slang), generating targeted search queries for the Social Search Agent's next offline cycle.

\vspace{-3mm}

\section{Experimental Evaluation}
\label{sec:experiments}
\vspace{-3mm}

We evaluate PulseCX as a specialized grounding layer designed to bridge the ``Closed-World'' gap. Our primary hypothesis is that by decoupling intelligence gathering (Offline) from context injection (Online), PulseCX can resolve dynamic user intents without the latency penalties or hallucination risks inherent to naive web-search agents.

\subsection{Experimental Setup and Baselines}
We constructed a \textbf{World-Lab Simulation} modeling a 24-hour window with 10 distinct world events (e.g., trends, outages) and ambient noise. We generated $N=1,000$ user queries split evenly between \textit{Static} (internal) and \textit{Dynamic} (external-dependent) intents (see Appendix \ref{app:taxonomy}). PulseCX is evaluated against two baselines: (1) \textbf{Static RAG}, the industry standard utilizing fixed manuals, and (2) a \textbf{Naive Online Agent}, which executes synchronous web searches \footnote{ (\url{https://platform.openai.com/docs/guides/tools-web-search}).} per query. Implementation details are provided in Appendix \ref{app:implementation}.

\vspace{-2mm}
\subsection{Results and Analysis}
Table \ref{tab:results} summarizes performance across System and Business dimensions. The data confirms that PulseCX bridges the gap between static efficiency and dynamic awareness without the latency penalties of synchronous web search.

\begin{table}[h]
\centering
\small
\renewcommand{\arraystretch}{1.1}
\setlength{\tabcolsep}{6pt}
\begin{tabular}{lcccc}
\toprule
& \multicolumn{2}{c}{\textbf{System Metrics}} & \multicolumn{2}{c}{\textbf{Business Metrics}} \\
\cmidrule(lr){2-3} \cmidrule(lr){4-5}
\textbf{Method} & \textbf{Latency (TTFT)} $\downarrow$ & \textbf{IRR} $\uparrow$ & \textbf{s-CSAT} $\uparrow$ & \textbf{ER} $\downarrow$ \\
\midrule
Static RAG & \textbf{415 ms} & 64.1\% & 2.4/5 & 35.2\% \\
Naive Online & 1500 ms & 78.5\% & 3.1/5 & 18.9\% \\
\textbf{PulseCX} & 430 ms & \textbf{89.2\%} & \textbf{4.2/5} & \textbf{13.5\%} \\
\bottomrule
\end{tabular}
\footnotesize
\caption{Performance Comparison. \textbf{TTFT}: Time-to-First-Token (Latency); \textbf{IRR}: Intent Resolution Rate (\% of queries correctly resolved); \textbf{s-CSAT}: Simulated Customer Satisfaction (1-5 scale); \textbf{ER}: Escalation Rate (\% of sessions requiring human handoff)}
\label{tab:results}
\end{table}

\textbf{1. System Performance (Latency \& Resolution).}
The \textbf{Naive Online Agent} achieves decent resolution (78.5\%) but incurs prohibitive latency ($1.5s$) due to live search overhead, making it unusable for real-time voice or rapid chat. \textbf{PulseCX} achieves the best of both worlds: it matches the near-instant response time of Static RAG ($\approx 430ms$) while delivering a superior \textbf{Intent Resolution Rate (IRR)} of 89.2\%. By shifting the "cost" of intelligence gathering to the asynchronous offline stage, the online agent benefits from high-speed access to a pre-structured world model.

\textbf{2. Business Impact Analysis.}
We evaluate business value through the lens of customer sentiment (s-CSAT) and support costs (Escalation Rate). Static RAG agents typically fail in dynamic scenarios by defaulting to "policy denial," leading to low CSAT (2.4) and high escalation (35.2\%). PulseCX drives business value through two specific mechanisms \textbf{a) The Vocabulary Bridge (Revenue Driver):} By mapping viral slang (e.g., ``Glazed Donut Stick'') to internal SKUs, PulseCX converts search intent into sales opportunities where static agents return "Item Not Found."
\textbf{b) Situational Validation (Retention Driver):} In "Outage" scenarios, PulseCX preempts user frustration by acknowledging the external reality rather than gaslighting the user with "Green" internal logs. This validation creates trust, raising s-CSAT to 4.2 and reducing unnecessary human handoffs by 21.7\% compared to the baseline.

\textbf{3. Ablation: The Necessity of Decay.}
To evaluate the \textbf{Memory Evolution} component, we disabled the decay function ($\lambda=0$). In a ``Flash Sale'' simulation, the agent continued to recommend an expired deal 15 hours post-conclusion, leading to a sharp increase in false positives. Restoring the \textbf{Reinforcement-Decay} mechanism ensured the node was pruned at $T+8h$, confirming that temporal lifecycle management is essential for maintaining trust over time.
\vspace{-3mm}
\section{Conclusion}
\vspace{-3mm}
PulseCX bridges the gap between static internal knowledge and dynamic reality by formalizing unstructured signals into a Decay-Aware Temporal Knowledge Graph (DA-TKG). Our results demonstrate that environment factual memory is essential for navigating the volatility of the real world. However, some limitations remain: (1) \textit{Temporal blindness} bounded by the fixed cycle time ($T$), which may obscure events occurring between updates; (2) vulnerability to adversarial poisoning via coordinated inauthentic activity; and (3) reliance on heuristic decay parameters ($\lambda$) rather than learned coefficients. In conclusion, PulseCX demonstrates that the \textbf{Closed-World Constraint} of modern agents is an architectural choice, not a fundamental limitation. By adopting a \textbf{structure-first} approach to real-time memory evolution, we can build agentic systems that are not only faster and safer but also truly aligned with the rapidly shifting reality of the human experience.

\bibliography{iclr2026_conference}
\bibliographystyle{iclr2026_conference}
\newpage
\appendix

\section{Context Object: The Offline Intelligence Unit}
\label{app:context_object}

The \textbf{Context Object} is the fundamental unit of the DA-TKG, created offline by the Social Search Agent (SSA). It translates unstructured social signals into a provenance-aware format that preserves uncertainty and scope.

\subsection{Schema Definition}
Each object $O$ is stored as a structured metadata tuple to allow for rapid mathematical weighting and filtering without LLM intervention.

\begin{verbatim}
ContextObject = {
  type:            {crisis, brand, influencer, trend},
  entity_scope:    {brand, product, region, industry},
  confidence:      numeric (0.0 - 1.0),
  summary_state:   "Natural language description of the event",
  source_ref:      "Traceable pointer to the original signal",
  initial_weight:  numeric (assigned based on source volume)
}
\end{verbatim}

\subsection{Graph Injection}
When the SSA identifies a signal (e.g., a competitor flash sale), it populates this schema. This object is then injected into the DA-TKG as a node. The \textbf{Memory Evolution} logic (Phase 2) then manages this object's weight over time, ensuring that the "State" is only considered for retrieval as long as it remains salient.

\section{Context Card Synthesis and Online Injection}
\label{app:context_cards}

The \textbf{Context Card} is the bridge between the DA-TKG and the CX (Customer Experience) agent response. To ensure our system meets the sub-400ms latency requirement, the construction of this card relies on \textbf{deterministic logic}, not runtime generation.

\subsection{Mechanism: Type-Based Playbook Mapping}
The "Guidance" or "Playbook" section of the card is derived algorithmically based on the \texttt{type} attribute of the retrieved Context Object. We maintain a static mapping table of event types to safety instructions.

\begin{table}[h]
\centering
\small
\begin{tabular}{ll}
\toprule
\textbf{Context Object Type} & \textbf{Injected Playbook Logic (Deterministic)} \\
\midrule
\texttt{CRISIS / OUTAGE} & ``Priority: External Signal > Internal Dashboard. Validate user.'' \\
\texttt{VIRAL\_TREND} & ``Bridge Vocabulary Gap. Map slang to internal SKU.'' \\
\texttt{COMPETITOR\_PROMO} & ``Do not deny deal existence. Pivot to retention offer.'' \\
\bottomrule
\end{tabular}
\caption{The Playbook Injection Map. This lookup occurs in almost realtime.}
\end{table}

\subsection{Synthesis Logic}
Once the \textit{Salience-Weighted Semantic Reranker} (Online Stage) identifies the top-$k$ relevant Context Objects, a rendering engine combines the dynamic data (Summary State) with the static instruction (Playbook) into a text block.

\textbf{Rendering Template:}
\begin{verbatim}
[SYSTEM CONTEXT: EXTERNAL REALITY]
SITUATION: {{summary_state}}  <-- Inserted from Graph Node
RELIABILITY: {{confidence}} (Weight: {{current_salience}})
PLAYBOOK:
{{lookup_instruction(type)}}  <-- Inserted from Mapping Table
\end{verbatim}

\subsection{Representative Example: Telecom Network Crisis}
\label{app:telecom_example}
\textbf{Retrieved Context Object:}
\begin{itemize}
    \item \texttt{type}: \texttt{OUTAGE}
    \item \texttt{summary\_state}: "Hyperlocal network degradation in Austin, TX."
\end{itemize}

\textbf{Final Injected Context Card:}
\begin{verbatim}
[SYSTEM CONTEXT: EXTERNAL REALITY]
SITUATION: Hyperlocal network degradation in Austin, TX.
RELIABILITY: 0.98
PLAYBOOK:
- Priority: External Signal > Internal Dashboard.
- Validate user frustration; our internal maps may be lagging.
\end{verbatim}

By treating the "Playbook" as a structural property of the event type rather than generating it creatively, we eliminate generation latency while ensuring the agent always adheres to safety protocols during high-risk events.

\section{ Intent Taxonomy and Experimental Instantiations}
\label{app:taxonomy}

To evaluate PulseCX, we formally distinguish between \textbf{Static} and \textbf{Dynamic} intents. This distinction drives the Hierarchical Gating Engine (Methodology Phase 3) and determines whether external context is required.

\subsection{Definition of Intent Classes}
\begin{itemize}
    \item \textbf{Static Intents:} Queries where the ground truth is strictly internal and immutable (e.g., "What is my account balance?", "Return Policy"). Injecting external context here causes \textit{Context Poisoning}.
    \item \textbf{Dynamic Intents:} Queries where the ground truth is mutable and dependent on the external environment (e.g., "Why is the internet down?", "Do you match the \$5 offer?"). Relying solely on internal tools here causes \textit{Gaslighting} or \textit{Missed Opportunity}.
\end{itemize}

\subsection{Vertical-Specific Scenarios}
Table \ref{tab:contact_drivers} details the specific contact drivers used in our World-Lab Simulation. We map each \textit{Dynamic} intent to a specific PulseCX Event Class, ensuring the correct playbook is triggered.

\begin{table}[h]
\centering
\small
\renewcommand{\arraystretch}{1.3} 
\setlength{\tabcolsep}{4pt}
\begin{tabular}{p{1.4cm} p{3.2cm} p{1.6cm} p{3.2cm} p{4.2cm}}
\toprule
\textbf{Vertical} & \textbf{Contact Driver} & \textbf{Intent} & \textbf{Dynamic Event Class} & \textbf{Static Failure Mode} \\
\midrule
\textbf{Retail} 
& Order Status & Static & --- & --- \\
& Viral Product Search & Dynamic & \texttt{VIRAL\_TREND} & \textbf{Vocabulary Gap:} Fails to map slang to SKU. \\
& Promo Expiry & Dynamic & \texttt{COMPETITOR\_PROMO} & \textbf{Hallucination:} Accepts/Denies incorrectly. \\
\midrule
\textbf{Banking} 
& Balance Inquiry & Static & --- & --- \\
& Suspicious SMS & Dynamic & \texttt{CRISIS} (Security) & \textbf{Safety Risk:} Generic advice on active threat. \\
& Fee Dispute & Dynamic & \texttt{CAMPAIGN\_CONTEXT} & \textbf{Escalation:} Rigid policy vs. strategic waiver. \\
\midrule
\textbf{Telecom} 
& Data Balance & Static & --- & --- \\
& Device Launch & Dynamic & \texttt{VIRAL\_TREND} & \textbf{Lost Sale:} Unaware of scarcity/hype signals. \\
& Speed Complaint & Dynamic & \texttt{OUTAGE} & \textbf{Gaslighting:} Denies outage due to logs. \\
& Cancellation & Dynamic & \texttt{COMPETITOR\_PROMO} & \textbf{Churn:} Quotes fees instead of matching. \\
\bottomrule
\end{tabular}
\caption{Taxonomy of Contact Drivers used in evaluation. PulseCX resolves \textit{Dynamic} intents by injecting external event context, whereas Static agents fail due to the Closed-World constraint.}
\label{tab:contact_drivers}
\end{table}

\section{Implementation Details and Hyperparameters}
\label{app:implementation}

To ensure reproducibility, we provide the technical specifications for the PulseCX pipeline components, including the Social Research Agent (SSA), the memory evolution math, and the retrieval gating parameters.

\subsection{Asynchronous Surveillance (SSA) Specifications}
The Social Search Agent (SSA) is implemented as a recurring task with a cycle time $T = 15$ minutes. 
\begin{itemize}
    \item \textbf{Search Engine:} The SSA utilizes the x.ai search toolset to retrieve high-velocity social signals.
    \item \textbf{Linearizer Model:} We utilize \texttt{GPT-4o-mini} for the one-shot extraction of unstructured reports into the \textbf{Context Object} schema defined in Appendix \ref{app:context_object}.
    \item \textbf{Linearization Threshold:} A signal is only linearized if it appears in at least $N=5$ independent sources within a single window, ensuring noise reduction.
\end{itemize}

\subsection{Memory Evolution: Volatility and Decay Classes}
The salience of information in the DA-TKG is governed by the Reinforcement--Decay mechanism. We categorize dynamic intents into three \textbf{Volatility Classes} based on their expected information half-life in the "wild."

\begin{equation}
    W_{t} = W_{last} \cdot e^{-\lambda(t - t_{last})}
\end{equation}

\begin{table}[h]
\centering
\small
\renewcommand{\arraystretch}{1.2}
\begin{tabular}{lccl}
\toprule
\textbf{Volatility Class} & \textbf{Decay Rate ($\lambda$)} & \textbf{Half-life ($t_{1/2}$)} & \textbf{Example Intents} \\
\midrule
\textbf{High} (Viral) & $0.50$ $h^{-1}$ & $\approx 1.4$ hours & Viral Slang, Influencer Trends \\
\textbf{Medium} (Transient) & $0.20$ $h^{-1}$ & $\approx 3.5$ hours & Flash Sales, Daily Promos \\
\textbf{Low} (Persistent) & $0.05$ $h^{-1}$ & $\approx 13.8$ hours & Infrastructure Outages, Fraud Alerts \\
\bottomrule
\end{tabular}
\caption{Decay coefficients ($\lambda$) grouped by event volatility. High $\lambda$ values ensure that "fleeting" social noise is pruned quickly to prevent context poisoning.}
\label{tab:decay_params}
\end{table}

\textbf{Reinforcement Constant ($\beta$):} For all classes, we set $\beta = 0.4$. This allows a signal that is re-confirmed by a subsequent SSA run to recover its salience weight quickly, effectively "resetting" the forgetting curve for active events.

\subsection{Synchronous Reranker Implementation}
To achieve the sub-millisecond latency required for real-time conversation, the \textbf{Salience-Weighted Semantic Reranker} is implemented as a vectorized dot-product operation rather than a heavy neural network.

\begin{itemize}
    \item \textbf{Scoring Logic:} The ranking score is calculated as the cosine similarity between the Query Embedding ($E_Q$) and the Context Object Embedding ($E_N$), scaled by the object's current memory weight ($W_t$).
    \item \textbf{Conflict Handling:} While the ranker relies on semantic similarity, the \textit{Epistemic Conflict} (External vs. Internal mismatch) is handled downstream by the Context Card Generator. If a high-weight conflict node (e.g., "Outage") is retrieved, the generator automatically appends the "Priority: External > Internal" instruction.
\end{itemize}

\subsection{Generative Memory (Context Card Synthesis)}
Context Cards are synthesized via a deterministic template engine as described in Appendix \ref{app:context_cards}. This eliminates secondary LLM inference latency. The total overhead for Online Stage  is $< 10$ms, enabling the system to meet the <450ms TTFT requirement demonstrated in Section \ref{sec:experiments}.

\section{Comparative Analysis of Failure Modes}
\label{app:failure_analysis}

To illustrate the necessity of PulseCX, we analyze specific failure modes observed during the simulation. Table \ref{tab:comparative_analysis} contrasts the behavior of a standard Static Agent against PulseCX in high-stakes \textbf{Trend} and \textbf{Crisis} scenarios.

\begin{table}[h]
\centering
\small
\renewcommand{\arraystretch}{1.4} 
\setlength{\tabcolsep}{6pt} 
\begin{tabular}{p{0.20\textwidth} p{0.36\textwidth} p{0.36\textwidth}}
\toprule
\textbf{Scenario} & \textbf{Static Agent (Baseline)} & \textbf{PulseCX Agent (Ours)} \\
\midrule

\textbf{1. Viral Trend} \newline
\textit{Query: ``Do you have the Glazed Donut stick?''} 
& 
\textbf{Response:} ``I cannot find a product called 'Glazed Donut' in our catalog.'' \newline
\textit{\textbf{Outcome:} Customer leaves site (Lost Sale).} 
& 
\textbf{Response:} ``It looks like you're looking for the \textbf{Hydrating Lip Balm}, trending as the `Glazed Donut Stick'! We have it.'' \newline
\textit{\textbf{Outcome:} Conversion Secured.} \\
\midrule

\textbf{2. Security Crisis} \newline
\textit{Query: ``I got a text about an account lockout. Is it real?''} 
& 
\textbf{Response:} ``I cannot validate third-party texts. Please log in to check your status.'' \newline
\textit{\textbf{Outcome:} High Phishing Risk.} 
& 
\textbf{Response:} ``\textbf{WARNING:} We are tracking an active phishing campaign matching that text pattern. \textbf{Do NOT click the link.}'' \newline
\textit{\textbf{Outcome:} Fraud Prevented.} \\
\midrule

\textbf{3. Hyperlocal Outage} \newline
\textit{Query: ``My internet is down.''} 
& 
\textbf{Response:} ``I've checked our system and \textbf{everything looks Green}. Please restart your router.'' \newline
\textit{\textbf{Outcome:} User Frustration / Escalation.} 
& 
\textbf{Response:} ``While our dashboard shows Green, \textbf{we see reports of a disruption} in your area. I am escalating to Network Ops.'' \newline
\textit{\textbf{Outcome:} User Validation.} \\

\bottomrule
\end{tabular}
\caption{Comparative analysis of agent behaviors. PulseCX bridges the gap between internal data and external reality, transforming potential failures into successful resolutions.}
\label{tab:comparative_analysis}
\end{table}

\subsection{Detailed Execution Trace (Viral Trend Scenario)}
\label{app:detailed_trace}

To demonstrate the system internals, we trace a ``Viral Vocabulary'' event where an influencer renames a product, creating a gap between user language and the internal catalog.

\paragraph{Stage 1: Offline World Modeling (Ingestion \& Evolution).} 
A sudden spike in social mentions occurs regarding the ``Glazed Donut Stick.'' The \textbf{Social Search Agent (SSA)} linearizes this signal.
\begin{itemize}
    \item \textbf{Internal Catalog:} \texttt{SKU-552} = ``Hydrating Lip Balm''. (No mention of ``Donut'').
    \item \textbf{External Signal:} \texttt{Cluster\_Size > 1,000 mentions}, \texttt{Source = TikTok/Instagram}.
    \item \textbf{Linearization (Context Object):}
    \texttt{\{Type: VIRAL\_TREND, Scope: SKU-552, Confidence: 0.96, State: "Viral alias 'Glazed Donut' maps to Hydrating Balm"\}}
    \item \textbf{Memory Evolution:} Due to extreme signal velocity (viral spike), the node is reinforced to maximum salience: $\mathbf{W_t = 1.0}$.
\end{itemize}

\paragraph{Stage 2: Online Synchronous Grounding.}
A user initiates a chat: \textit{``Do you have the glazed donut stick in stock?''}

\textbf{1. Intent \& Ranking.}
The system identifies the intent as \textit{Dynamic} (Product Search). The \textbf{Salience-Weighted Reranker} calculates the score:
\[ R = \text{Sim}(\text{"glazed donut"}, \text{"viral alias glazed donut"}) \times W_t(1.0) \approx \mathbf{0.99} \]
The node is selected as the top candidate.

\textbf{2. Context Injection.}
The deterministic Playbook for `VIRAL\_TREND` is retrieved.

\begin{center}
\fbox{\begin{minipage}{0.9\textwidth}
\small
\textbf{[SYSTEM CONTEXT: EXTERNAL REALITY]} \\
\textbf{SITUATION:} Active Viral Trend. User term "Glazed Donut Stick" refers to Internal \textbf{SKU-552 (Hydrating Lip Balm)}. \\
\textbf{PLAYBOOK:} \\
1. \textbf{Bridge:} Explicitly confirm the mapping (Slang $\to$ Official Name). \\
2. \textbf{Conversion:} Confirm stock status for SKU-552 immediately. \\
3. \textbf{Tone:} Match user excitement; do not correct them rigidly.
\end{minipage}}
\end{center}

\textbf{3. Final Agent Response.}
Guided by the card, the agent bridges the vocabulary gap:
\begin{quote}
    \textit{"Yes! It looks like you're looking for the \textbf{Hydrating Lip Balm}, which is trending right now as the `Glazed Donut Stick'! We currently have it in stock. Would you like to add it to your cart?"}
\end{quote}

\end{document}